\pdfoutput=1
\documentclass[11pt,a4paper]{article}
\usepackage[hyperref]{acl2020}
\usepackage{times}
\usepackage{latexsym}

\usepackage{microtype}

\aclfinalcopy 
\def\aclpaperid{2184} 

\usepackage[nameinlink,capitalize]{cleveref}  
\crefname{equation}{}{}
\crefformat{section}{#2\S#1#3}
\crefformat{subsection}{#2\S#1#3}
\crefmultiformat{section}{#2\S#1#3}{ and~#2\S#1#3}{, #2\S#1#3}{, and~#2\S#1#3}
\crefmultiformat{subsection}{#2\S#1#3}{ and~#2\S#1#3}{, #2\S#1#3}{, and~#2\S#1#3}
\crefrangeformat{section}{#3\S#1#4--#5#2#6}
\crefrangeformat{subsection}{#3\S#1#4--#5#2#6}
\newcommand{\seeref}[1]{(see~\cref{#1})}
\newcommand{\citex}[3][]{\citep[#2:][#1]{#3}}  
\newcommand{\citeg}[2][]{\citex[#1]{for example}{#2}}  

\newcommand{\sortbib}[1]{}

\title{What are the Goals of Distributional Semantics?}

\author{Guy Emerson \\
	Department of Computer Science and Technology \\
	University of Cambridge \\
	\texttt{gete2@cam.ac.uk}}

\date{}

\begin{document}
\maketitle
\begin{abstract}

Distributional semantic models have become a mainstay in NLP,
providing useful features for downstream tasks.
However, assessing long-term progress requires explicit long-term goals.
In this paper, I take a broad linguistic perspective,
looking at how well current models can deal with various semantic challenges.
Given stark differences between models proposed in different subfields,
a broad perspective is needed to see how we could integrate them.
I conclude that, while linguistic insights can guide the design of model architectures,
future progress will require balancing the often conflicting demands of
linguistic expressiveness and computational tractability.

\end{abstract}

\section{Introduction}

In order to assess progress in any field,
the goals need to be clear.
In assessing progress in semantics,
\citet{koller2016top} contrasts
``top-down'' and ``bottom-up'' approaches:
a top-down approach begins with an overarching goal,
and tries to build a model to reach it;
a bottom-up approach begins with existing models,
and tries to extend them towards new goals.\footnote{%
	For further discussion, see: \citet{bender2020understand}.
}
Like much of NLP, distributional semantics is largely bottom-up:
the goals are usually to improve performance on particular tasks,
or particular datasets.
Aiming to improve NLP applications is of course a legitimate decision,
but \citeauthor{koller2016top} points out a problem if there is no top-down goal:
``Bottom-up theories are intrinsically unfalsifiable...
We won't know where distributional semantics is going until it has a top-down element''.
This is contrasted against truth-conditional semantics,
a traditional linguistic approach which is largely top-down:
``truth-conditional semantics hasn't reached its goal,
but at least we knew what the goal was''.

In this paper, I take a long-term linguistic perspective,
where the top-down goal is to characterise
the meanings of all utterances in a language.
This is an ambitious goal, and a broad one.
To make this goal more precise,
in the following sections I will elaborate on several aspects of meaning
which could be considered crucial.
For each aspect, I identify a plausible goal,
lay out out the space of possible models,
place existing work in this space,
and evaluate which approaches seem most promising.
By making the goals explicit,
we can assess whether we are heading in the right direction,
and we can assess what still needs to be done.
If a reader should disagree with my conclusions,
they should start by looking at my goals.


\section{Background: Distributional Semantics}
\label{sec:back}

The aim of distributional semantics is to learn the meanings of linguistic expressions
from a corpus of text.
The core idea, known as the \textit{distributional hypothesis},
is that the contexts in which an expression appears
give us information about its meaning.\footnote{%
	The hypothesis is often stated more narrowly,
	to say that similar words appear in similar contexts,
	but in this paper I am interested in semantics beyond just similarity.
}

The idea has roots in American structuralism \citep{harris1954distribution}
and British lexicology \citep{firth1951collocation,firth1957company}\footnote{%
	\citeauthor{firth1951collocation} used the term \textit{collocational}, not \textit{distributional}.
},
and with the advent of modern computing,
it began to be used in practice.
In a notable early work, \citet{spaerck-jones1964synonymy}
represented word meanings as boolean vectors,
based on a thesaurus.


Distributional semantics has become widespread in NLP,
first with the rise of count vectors
\citex{for an overview, see}{erk2012vector,clark2015vector},
then of word embeddings \citep{mikolov2013vector},
and most recently, of contextualised embeddings
\citep{peters2018elmo,devlin2019bert}.\footnote{%
	For connections between count vectors and embeddings,
	see: \citet{levy2014equiv,cotterell2017equiv};
	for connections with contextual embeddings:
	\citet{kong2020represent}.
}
What all of these approaches share
is that they learn representations
in an unsupervised manner on a corpus.

While much work takes a bottom-up approach,
as \citeauthor{koller2016top} observes,
a notable exception is the type-driven tensorial framework
of \citet{coecke2010tensor} and \citet{baroni2014tensor},
which has broad linguistic goals,
and will be mentioned in several sections below.
This framework represents the meanings of words as tensors,
and constructs phrase meanings using tensor contraction
based on predicate-argument structure.
For example, there is one vector space for nouns,
and a second vector space for sentences,
so intransitive verbs are matrices
(mapping noun vectors to sentence vectors).


\section{Meaning and the World}
\label{sec:world}

Language is always \textit{about} something.
In this section, I discuss challenges
in connecting a semantic model to things in the world.

\subsection{Grounding}
\label{sec:world:ground}

As \citet{harnad1990ground} discusses,
if the meanings of words are defined only in terms of other words,
these definitions are circular.
One goal for a semantic model is to capture how language relates to the world,
including sensory perception and motor control --
this process of connecting language to the world is called \textit{grounding}.\footnote{%
	This includes connecting abstract concepts to the world,
	although such connections are necessarily more indirect.
	For further discussion, see:
	\citet{blondinmasse2008dictionary,pecher2011abstract,pulvermueller2013abstract,barsalou2018abstract}
}

A purely distributional model is not grounded,
as it is only trained on text,
with no direct link to the world.
There are several ways we could try to ground a distributional model
\citex{for an overview, see}{baroni2016ground}.
The simplest way is to train a distributional model as normal,
then combine it with a grounded model.
For example, \citet{bruni2011visual}
concatenate distributional vectors and image feature vectors.
This has also been applied to other senses:
\citet{kiela2015olfactory} use olfactory data,
and \citet{kiela2017auditory} use both visual and auditory data.
However, while there is grounded information in the sensory dimensions,
concatenation leaves the distributional dimensions ungrounded.

A second approach is to find correlations between distributional and sensory features.
For example, \citet{bruni2014men} perform SVD on concatenated vectors,
\citet{silberer2014visual} train an autoencoder on concatenated vectors,
and \citet{lazaridou2014wampimuk} and \citet{bulat2016map}
learn a mapping from distributional vectors to visual vectors (and vice versa).
However, there is no guarantee that
every distributional feature will correlate with sensory features.
Distributional features without correlations will remain ungrounded.


Finally, a third approach is joint learning
-- we define a single model,
whose parameters are learnt based on both corpus data and grounded data.
For example, \citet{feng2010visual} train an LDA model \citep{blei2003lda}
for both words and ``visual words'' (clusters of visual features).
\citet{lazaridou2015ground} use a Skip-gram model \citep{mikolov2013vector}
to jointly predict both words and images.
\citet{kiros2014ground} embed both text and images in a single space,
training an RNN to process captions,
and a CNN to process images.
Pure distributional models look for word co-occurrence patterns,
while joint models prefer co-occurrence patterns that match the grounded data.
For this reason,
I believe joint learning is the right approach to ground corpus data --
semantic representations can be connected to grounded data from the outset,
rather than trying to make such connections after the fact.

However, we must still make sure that all distributional features are grounded.
With \citeauthor{feng2010visual}'s LDA model,
some topics might only generate words rather than ``visual words''.
Similarly, with \citeauthor{lazaridou2015ground}'s joint Skip-gram model,
some embeddings might only predict words rather than images.
Conversely, we also need to make sure that we make full use of corpus data,
rather than discarding what is difficult to ground.
For example, \citeauthor{kiros2014ground}'s joint embedding model
learns sentence embeddings in order to match them to images.
It is not obvious how this approach could be extended so that
we can learn embeddings for sentences that cannot be easily depicted in an image.

This leads to the question:
how should a joint architecture be designed,
so that we can fully learn from corpus data,
while ensuring that representations are fully grounded?
Grounding is hard, and indeed \citet{kuhnle2018quantifier} find that
some semantic constructions (such as superlatives)
are much harder for grounded models to learn than others.
In the following section, I discuss how language relates to the world.
Clarifying this relationship should help us to design good joint architectures.

\subsection{Concepts and Referents}
\label{sec:world:concept}

How do meanings relate to the world?
In truth-conditional semantics, the answer is that
meaning is defined in terms of \textit{truth}.\footnote{%
	For a discussion of this point, see: \citet{lewis1970sem}.
	For an\vspace{\maxdimen} 
	introduction to truth-conditional semantics, see:
	\citet{cann1993sem,allan2001sem,kamp2013sem}.
}
If an agent understands a language,
then in any given situation,
they know how to evaluate whether a sentence is true or false of that situation.\footnote{%
	On the notion of \textit{situation}, see:
	\citet{barwise1983situation}.
	On knowing \textit{how} to evaluate truth values
	vs.\ actually evaluating truth values, see:
	\citet{dummett1976meaning,dummett1978know}.
}
An advantage of this approach is that it supports logical reasoning,
which I will discuss in~\cref{sec:sentence:logic}.
One goal for a semantic theory is to be able to generalise to new situations.
This is difficult for traditional truth-conditional semantics,
with classical theories challenged on both
philosophical grounds \citeg[\S66--71]{wittgenstein1953investigations}
and empirical grounds \citeg{rosch1975concept,rosch1978categorize}.
However, a machine learning approach seems promising,
since generalising to new data is a central aim of machine learning.

For a semantic model to be compatible with truth-conditional semantics,
it is necessary to distinguish
a \textit{concept} (the meaning of a word)
from a \textit{referent} (an entity the word can refer to).\footnote{%
	Following \citet[pp.~4--5]{murphy2002concept},
	I use the term \textit{concept} without committing to a particular theory of concepts.
}
The importance of this distinction has been noted for some time
\citeg{ogden1923triangle}.
A concept's set of referents is called its \textit{extension}.\footnote{%
	Or \textit{denotation}.
	In psychology, the term \textit{category} is also used
	\citeg{smith1981concept,murphy2002concept}.
}

Even if we can construct grounded concept vectors,
as discussed in~\cref{sec:world:ground},
there is still the question of how to relate a concept vector to its referents.\footnote{%
	While distributional representations can be learnt for named entities
	\citeg{herbelot2015name,boleda2017instance},
	most real-world entities are not mentioned in text.
}
One option is to embed both concepts and entities in the same space.
We then need a way to decide how close the vectors need to be,
for the entity to be in the concept's extension.
A second option is to embed concepts and referents in distinct spaces.
We then need a way to relate the two spaces.

In both cases, we need additional structure beyond
representing concepts and referents as points.
One solution is to represent a concept by a \textit{region} of space
\citep{gaerdenfors2000space,gaerdenfors2014space}.
Entities embedded inside the region are referents,
while those outside are not.
For example, \citet{mcmahan2015colour} learn representations of colour terms,
which are grounded in a well-understood perceptual space.


A related idea is to represent a concept as a binary classifier,
where an entity is the input.\footnote{%
	For deterministic regions and classifiers,
	there is a one-to-one mapping between them,
	but this is not true for probabilistic regions and classifiers,
	due to covariance.
}
One class is the concept's extension,
and the other class is everything else.
\citet{larsson2013classifier} represents the meaning of a perceptual concept
as a classifier of perceptual input.
A number of authors have trained image classifiers using captioned images
\citeg{schlangen2016classifier,%
	zarriess2017classifier,%
	zarriess2017classifier2,%
	utescher2019classifier,%
	matsson2019classifier%
}.

Such representations have however seen limited use in distributional semantics.
\citet{erk2009region1,erk2009region2} and \citet{dong2018imposing} learn regions,
but relying on pre-trained vectors,
which may have already lost referential information (such as co-reference)
that we would like to capture.
\citet{jameel2017region} learn a hybrid model,
where each word is represented by a point (as a target word)
and a region (as a context word).
In my own work, I have learnt classifiers \citep{emerson2016,emerson2017a,emerson2017b},
but with a computationally expensive model that is difficult to train.
The computational challenge is partially resolved in my most recent work
\citep{emerson2020pixie},
but there is still work to be done in scaling up the model
to make full use of the corpus data.
The best way to design such a model,
so that it can both make full use of the data
and can be trained efficiently,
is an open question.

\section{Lexical Meaning}
\label{sec:lex}

In this section, I discuss challenges in representing the meanings of individual words.

\subsection{Vagueness}
\label{sec:lex:vague}

Entities often fall along a continuum
without a sharp cutoff between concepts.
This is called \textit{vagueness} (or \textit{gradedness}).
(For an overview, see:
\citealp{sutton2013prob}, chapter 1; \citealp{vandeemter2010vague}.)
For example, \citet{labov1973cup} investigated the boundaries
between concepts like \textit{cup}, \textit{mug}, and \textit{bowl},
asking participants to name drawings of objects. 
For typical referents, terms were used consistently;
meanwhile, for objects that were intermediate between concepts
(for example, something wide for a cup but narrow for a bowl),
terms were used inconsistently.
For these borderline cases,
a single person may make different judgements at different times
\citep{mccloskey1978judge}.
One goal for a semantic model is
to capture how it can be unclear whether an entity is an referent of a concept.



One approach is to use \textit{fuzzy} truth values,
which are not binary true/false,
but rather values in the range~[0,1],
where 0~is definitely false,
1~is definitely true,
and intermediate values represent borderline cases
\citep{zadeh1965fuzzy,zadeh1975fuzzy}.
Fuzzy logic has not seen much use in computational linguistics.\footnote{%
	\citet{carvalho2012fuzzy} survey fuzzy logic in NLP,
	noting that its use is in decline,
	but they do not mention distributional semantics.
	Proposals such as Monte Carlo Semantics \citep{bergmair2010fuzzy}
	and Fuzzy Natural Logic \citep{novak2017fuzzy}
	do not provide an approach to distributional semantics.
	A rare exception is \citet{runkler2016fuzzy},
	who infers fuzzy membership functions from pre-trained vectors.
}

A second solution is to stick with binary truth values,
but using probability theory to formalise uncertainty about truth,
as has been proposed in formal semantics
\citeg{lassiter2011vague,
	fernandez2014vague,
	sutton2015prob,
	sutton2017prob}.
%
At the level of a single concept,
there is not much to decide between
fuzzy and probabilistic accounts,
since both assign values in the range~[0,1].
However, we will see in~\cref{sec:sentence:logic}
that they behave differently at the level of sentences.

Uncertainty has also been incorporated into distributional vector space models.
\citet{vilnis2014gauss} extend \citeauthor{mikolov2013vector}'s Skip-gram model,
representing meanings as Gaussian distributions over vectors.
\citet{barkan2017bayes} incorporate uncertainty into Skip-gram using Bayesian inference
-- rather than optimising word vectors,
the aim is to calculate the posterior distribution over word vectors,
given the observed data.
The posterior is approximated as a Gaussian,
so these two approaches produce the same kind of object.
\citet{balkir2014mixed}, working within the type-driven tensorial framework
\seeref{sec:back},
uses a quantum mechanical ``mixed state" to model uncertainty in a tensor.
For example, this replaces vectors by matrices,
and replaces matrices by fourth-order tensors.

While these approaches represent uncertainty,
it is challenging to use them to capture vagueness.
This basic problem is this:
a distribution allows us to \textit{generate} referents of a concept,
but how can we go in the other direction,
to \textit{recognise} referents of a concept?
It is tempting to classify a point using the probability density at that point,
but if we compare a more general term
with a more specific term (like \textit{animal} and \textit{dog}),
we find a problem:
a more general term has its probability mass spread more thinly,
and hence has a lower probability density than the more specific term,
even if both terms could be considered true.
I argued in~\cref{sec:world:concept} that,
to talk about truth,
we need to represent predicates as regions of space or as classifiers.
While a distribution over a space might at first sight look like a region of space,
normalising the probability mass to sum to~1 makes
a distribution a different kind of object.

\subsection{Polysemy}
\label{sec:lex:poly}

The meaning of a word can often be broken up into distinct \textit{senses}.
Related senses are called \textit{polysemous}:
for example, \textit{school} can refer to a building or an institution.
In contrast, \textit{homonymous} senses are unrelated:
for example, a \textit{school} of fish.
All of the above senses of \textit{school} are also \textit{lexicalised}
-- established uses that a speaker would have committed to memory,
rather than inferring from context.
I will discuss context-dependent meaning in~\cref{sec:sentence:context},
and focus here on lexicalised meaning.
One goal for a semantic model
is to capture how a word can have a range of polysemous senses.

One solution is to learn a separate representation for each sense
(for example: \citealp{schuetze1998sense,rapp2004sense,li2015sense};
for a survey, see: \citealp{camachocollados2018sense}).
However, deciding on a discrete set of senses is difficult,
and practical efforts at compiling dictionaries have not provided a solution.
Indeed, the lexicographer Sue Atkins bluntly stated,
``I don't believe in word senses''.\footnote{%
	\citet{kilgarriff1997sense} and \citet{hanks2000sense} both quote Atkins.
}
Although the sense of a word varies across usages,
there are many ways that we could cluster usages into a discrete set of senses,
a point made by many authors
\citeg{spaerck-jones1964synonymy,
	kilgarriff1997sense,
	kilgarriff2007sense,
	hanks2000sense,
	erk2010sense}.
To quantify this intuition, \citet{erk2009sense,erk2013sense}
produced the WSsim and Usim datasets,
where annotators judged the similarity between dictionary senses,
and the similarity between individual usages, respectively.
\citet{mccarthy2016cluster} quantify ``clusterability'' in USim,
showing that for some words,
usages cannot be clustered into discrete senses.
A good semantic model should therefore be able to capture variation in meaning
without resorting to finite sense inventories.

We could instead learn a single representation for all polysemous senses together.
Indeed, \citet{ruhl1989monosemy} argues that even
frequent terms with many apparent senses,
such as \textit{bear} and \textit{hit},
can be analysed as having a single underspecified meaning,
with the apparent diversity of senses explainable from context.
The challenge is then to represent such a meaning
without overgeneralising to cases where the word wouldn't be used,
and to model how meanings are specialised in context.
The second half of this challenge will be discussed in~\cref{sec:sentence:context}.

I have already argued in previous sections that we should
move away from representing each word as a single vector.
As discussed in~\cref{sec:lex:vague},
words can be represented with distributions,
and such an approach has also been applied to modelling word senses.
For example, \citet{athiwaratkun2017multimodal} use a mixture of Gaussians,
extending \citeauthor{vilnis2014gauss}'s model to allow multiple senses.
However, this ultimately models a fixed number of senses
(one for each Gaussian).
In principle, a distribution could be parametrised in a more general way,
moving beyond finite mixture models.
In the type-driven tensorial framework \seeref{sec:back},
\citet{piedeleu2015mixed} use mixed quantum states,
similarly to \citeauthor{balkir2014mixed}'s approach \seeref{sec:lex:vague}.
Although they only propose this approach for homonymy,
it could plausibly be used for polysemy as well.

If a word is represented by a region, or by a classifier,
we don't have the problem of finite sense inventories,
as long as the region or classifier
is parametrised in a general enough way
-- for example, a multi-layer neural net classifier,
rather than a finite mixture of simple classifiers.

\subsection{Hyponymy}
\label{sec:lex:hyp}

In the previous two sections, I discussed meanings of single words.
However, words do not exist on their own,
and one goal for semantic model is to represent relations between them.
A classic relation is \textit{hyponymy},\footnote{%
	This is also referred to as \textit{lexical entailment},
	making a link with logic \seeref{sec:sentence:logic}.
	Other relations include antonymy, meronymy, and selectional preferences.
	For reasons of space, I have decided to discuss one relation in detail,
	rather than many relations briefly.
	Hyponymy could be considered basic.
}
which describes when one term (the \textit{hyperonym} or \textit{hypernym})
has a more general meaning than another (the \textit{hyponym}).
Words that share a hyperonym are called \textit{co-hyponyms}.

In a vector space model,
it is not clear how to say if one vector is more general than another.
One idea is that a hyperonym should occur in all the contexts of its hyponyms.
This is known as the \textit{Distributional Inclusion Hypothesis}
(DIH; \citealp{weeds2004similarity,geffet2005inclusion}).
Using this idea and tools from information retrieval, \citet{kotlerman2009balapinc,kotlerman2010balapinc}
define the ``balAPinc'' measure of hyponymy.
\citet{herbelot2013hyponym} view a vector as a distribution over contexts,
using KL-divergence to measure hyponymy.
\citet{rei2013phd} gives an overview of hyponymy measures,
and proposes a weighted cosine measure.
For embeddings, the motivation for such measures is less direct,
but dimensions can be seen as combinations of contexts.
Indeed, \citet{rei2014hyponym} find
embeddings perform almost as well as count vectors.

However, a speaker is likely to choose an expression
with a degree of generality appropriate for the discourse
(the Maxim of Quantity; \citealp{grice1967maxim}),
and hence the DIH can be questioned.
\citet{rimell2014entail} points out that some contexts are highly specific.
For example, \textit{mane} is a likely context of \textit{lion}
but not \textit{animal},
even though \textit{lion} is a hyponym of \textit{animal},
contradicting the DIH.
\citeauthor{rimell2014entail} instead proposes measuring hyponymy using \textit{coherence}
(formalised using pointwise mutual information):
the contexts of a general term minus those of a hyponym are coherent,
but the reverse is not true.

Moving away from count vectors and pre-trained embeddings, there are other options.
One is to build the hyponymy relation into the definition of the space.
For example, \citet{vendrov2016order} use non-negative vectors,
where one vector is a hyponym of another
if it has a larger value in every dimension.
They train a model on WordNet \citep{miller1995wordnet,fellbaum1998wordnet}.
Building on this, \citet{li2017order} learn from both WordNet and text.

However, for a hierarchy like WordNet,
there are exponentially more words lower down.
This cannot be embedded in Euclidean space
without words lower in the hierarchy being increasingly close together.
\citet{nickel2017hyperbolic} propose using hyperbolic space,
where volume increases exponentially as we move away from any point.
\citet{tifrea2019poincare} build on this,
adapting Glove \citep{pennington2014glove}
to learn hyperbolic embeddings from text.
However, this approach does not generalise to non-tree hierarchies
-- for example, WordNet gives \textit{bass}
as a hyponym of \textit{singer}, \textit{voice}, \textit{melody}, \textit{pitch}, and \textit{instrument}.
Requiring that \textit{bass} is represented close to all its hyperonyms
also forces them close together (by the triangle inequality),
which we may not want, since they are in distant parts of the hierarchy.

Alternatively, we can view hyponymy as classification,
and simply use distributional vectors to provide input features
\citeg{weeds2014hyponym,rei2018hyponym}.
However, under this view, hyponymy is an opaque relationship,
making it difficult to analyse why one vector is classified as a hyponym of another.
Indeed, \citet{levy2015hyponym} find that such classifiers
mainly learn which words are common hyperonyms.

Moving away from vector representations,
it can be easier to define hyponymy.
\citet{erk2009region1,erk2009region2} and \citet[\S6.4]{gaerdenfors2014space}
discuss how using regions of space provides a natural definition:
$P$~is a hyponym of~$Q$
if the region for~$P$ is contained in the region for~$Q$.
\citet{bouraoui2017space} and \citet{vilnis2018box}
use this idea for knowledge base completion,
and \citet{bouraoui2020neighbor} build on this,
using corpus data to identify ``conceptual neighbours''.
In the type-driven tensorial framework \seeref{sec:back},
\citet{bankova2019hyponym} and \citet{lewis2019hyponym}
model words as normalised positive operators,
with hyponymy defined in terms of subspaces (eigenspaces).

Probability distributions also allow us to define hyponymy,
but it is harder than for regions,
since a distribution over a smaller region has higher probability density.
\citet{vilnis2014gauss} propose using KL-divergence.
\citet{athiwaratkun2018hyponym} propose a thresholded KL-divergence.
In the type-driven tensorial framework,
\citet{balkir2014mixed} proposes using a quantum version of KL-divergence,
which can be extended to phrases
\citep{balkir2015sentence,sadrzadeh2018sentence}.

However, detecting hyponymy from corpus data remains challenging.
Even in recent shared tasks \citep{bordea2016semeval,camachocollados2018semeval},
many systems use pattern matching, following \citet{hearst1992hyponym}.
For example, a string of the form \textit{X such as Y}
suggests that \textit{Y} is a hyponym of \textit{X}.
In the above shared tasks,
the best performing systems did not rely solely on distributional vectors,
but used pattern matching as well.

Although much work remains to be done in developing learning algorithms
which can detect hyponymy,
I believe that a region-based approach is the most promising.
Not only does it give a simple definition,
but it is also motivated for other reasons,
discussed elsewhere in this paper.

\section{Sentence Meaning}
\label{sec:sentence}

In the previous section, I discussed meaning at the level of words.
I now turn to challenges in representing meaning at the level of sentences.

\subsection{Compositionality}
\label{sec:sentence:compose}

Language is \textit{productive} --
a fluent speaker can understand a completely new sentence,
as long as they know each word and each syntactic construction in the sentence.
One goal for a semantic model
is to be able to \textit{derive} the meaning of a sentence from its parts,
so it can generalise to new combinations.
This is known as \textit{compositionality}.\footnote{%
	\citet{kartsaklis2013disambiguate} discuss how
	composition is often conflated with \textit{disambiguation},
	since composing ambiguous expressions often disambiguates them.
	Disambiguation can be seen as a kind of
	\textit{contextualisation} or \textit{context dependence},
	which I discuss in~\cref{sec:sentence:context}.
	The focus in this section is on deriving semantic representations for larger expressions.
}



For vector space models,
the challenge is how to compose word vectors
to construct phrase representations.
If we represent both words and phrases in the same vector space,
the challenge is to find a composition function that maps a pair of vectors to a new vector.
In the general case, this must be sensitive to word order,
since changing word order can change meaning.
\citet{mitchell2008compose,mitchell2010compose} compare a variety of such functions,
but find that componentwise multiplication performs best,
despite being commutative, and hence insensitive to word order.
The effectiveness of componentwise multiplication and addition
has been replicated many times
\citeg{baroni2010matrix,blacoe2012compose,rimell2016relpron,czarnowska2019}.
However, it is unclear how to adapt it to take word order into account,
and \citet{polajnar2014long} show that performance degrades with sentence length.

Alternatively, we can use a sentence space distinct from the word space.
This is often done with a task-based perspective
-- words are combined into sentence representations,
which are useful for solving some task.
For example, the final state of an RNN can be seen
as a representation of the whole sequence.
To make the composition more linguistically informed,
the network can be defined to follow a tree structure, rather than linear order \citeg{socher2010recursive,socher2012recursive,tai2015treelstm},
or even to learn latent tree structure
\citeg{dyer2016rnng,maillard2018tree}.
Alternatively, a sequence of token representations
can be combined using attention, which calculates a weighted sum,
as in a Transformer architecture \citep{vaswani2017attention}.

Regardless of architecture, the model can be optimised either for a supervised task,
such as machine translation \citeg{cho2014translate},
or for an unsupervised objective,
as in an autoencoder \citeg{hermann2013compose}
or language model \citeg{peters2018elmo,devlin2019bert}.
If we take a task-based perspective,
it is difficult to know if the representations
will transfer to other tasks.
In fact, \citet{changpinyo2018multitask}
find that for some combinations of tasks,
training on one task can be harmful for another.

As an alternative to task-based approaches,
the tensorial framework mentioned in~\cref{sec:back}
also uses sentence vectors,
but using tensor contraction to compose representations
based on argument structure.\footnote{%
	\citet{zanzotto2015comp} show how
	sentence similarity in this framework
	decomposes in terms of similarity of corresponding parts,
	because composition and dot products are linear.
}
\citet{polajnar2015sentence} explore sentence spaces
with dimensions defined by co-occurrences.

However, a weakness with the above approaches
is that they map sentences to a finite-dimensional space.
As we increase sentence length,
the number of sentences with distinct meanings increases exponentially.
For example, consider relative clauses:
\textit{the dog chased the cat};
\textit{the dog chased the cat which caught the mouse}; and so on.
To keep these meanings distinct, we have two options.
If the meanings must be a certain distance apart,
the magnitudes of sentence vectors need
to increase exponentially with sentence length,
so there is enough space to distinguish them.\footnote{%
	This can be formalised information-theoretically.
	Consider sending a message as a $D$-dimensional vector, through a noisy channel.
	If there is an upper bound~$K$ to the vector's magnitude,
	the channel has a finite \textit{channel capacity}.
	The capacity scales as~$K^D$, which is only polynomial in~$K$.
}
Alternatively, if the meanings can be arbitrarily close,
we need to record each dimension to a high precision
in order to distinguish the meanings.
The fine-grained structure of the space then becomes important,
but small changes to model parameters (such as updates during training)
would cause drastic changes to this structure.
I do not know any work exploring either option.
Otherwise, we are forced to view sentence vectors as lossy compression.\footnote{%
	This conclusion has been drawn before
	\citeg[p.~370]{goodfellow2016deep},
	but my argument makes the conditions more precise.
}
As \citet{mooney2014cram} put it:
``You can't cram the meaning of a whole  
\%\&!\$\# sentence into a single \$\&!\#* vector!''

Although compression can be useful for many tasks,
full and detailed semantic representations also have their place.
This is particularly important at a discourse level:
it would be absurd to represent, as vectors of the same dimensionality,
both a five-word sentence and the whole English Wikipedia.
However, this leaves open the question of how we \textit{should} represent sentence meaning.
In the following section, I turn to logic as a guide.


\subsection{Logic}
\label{sec:sentence:logic}

Sentences can express complex thoughts,
and build chains of reasoning.
Logic formalises this,
and one goal for a semantic model is to support the logical notions
of \textit{truth} (discussed in~\cref{sec:world:concept}),
and \textit{entailment} (one proposition following from another).

Vectors do not have logical structure,
but can still be used to provide features for a logical system,
for example if entailment is framed as classification:
given a \textit{premise} and \textit{hypothesis}, 
the task is to decide if the premise entails the hypothesis, contradicts it, or neither.
Datasets include SNLI \citep{bowman2015snli} and MultiNLI \citep{williams2018mnli}.

However, it is difficult to analyse approaches that do not use an explicit logic.
In fact, \citet{gururangan2018artifact} suggest that
high performance may be due to annotation artifacts:
only using the hypothesis,
they achieve 67\% on SNLI and 53\% on MultiNLI,
much higher than the majority class baseline (34\% and 35\%, respectively).
Performance on such datasets
may therefore overestimate the ability of neural models to perform inference.

To explicitly represent logical structure, there are a few options.
One is to build a hybrid system, combining a vector space with a logic.
For example, \citet{herbelot2015quantifier} aim to give logical interpretations to vectors.
They consider a number of properties (such as: \textit{is\_edible}, \textit{has\_a\_handle}, \textit{made\_of\_wood}),
and for each, they learn a mapping from vectors to values in~$[0,1]$,
where 0 means the property applies to no referents,
and 1 means it applies to all referents.
This is an interesting way to probe what information is available in distributional vectors,
but it is unclear how it could be generalised to deal with individual referents
(rather than summarising them all),
or to deal with complex propositions
(rather than single properties).

\citet{garrette2011logic} and \citet{beltagy2016logic}
incorporate a vector space model into a Markov Logic Network \citep{richardson2006markov},
a kind of probability logic.
If two predicates have high distributional similarity,
they add a probabilistic inference rule saying that,
if one predicate is true of an entity, the other predicate is likely to also be true.
This allows us to use distributional vectors in a well-defined logical model,
but it assumes we can interpret similarity in terms of inference
\citex{for discussion, see}{erk2016alligator}.
As argued in \cref{sec:world} above,
pre-trained vectors may have already lost information,
and in the long term, it would be preferable to learn logical representations directly.

\citet{lewis2013logic} use a classical logic,
and cluster predicates that are observed to hold of the same pairs of named entities
-- for example, \textit{write}(\textit{Rowling}, \textit{Harry Potter})
and \textit{author}(\textit{Rowling}, \textit{Harry Potter}).
This uses corpus data directly, rather than pre-trained vectors.
However, it would need to be generalised
to learn from arbitrary sentences,
and not just those involving named entities.

A second option is to define a vector space with a logical interpretation.
\citet{grefenstette2013tensor} gives a logical interpretation
to the type-driven tensorial framework \seeref{sec:back},
where the sentence space models truth values,
and the noun space models a domain of $N$ entities.
However, \citeauthor{grefenstette2013tensor} shows that quantification
would be nonlinear,
so cannot be expressed using tensor contraction.
\citet{hedges2019quantifier} provide an alternative account
which can deal with quantifiers,
but at the expense of noun dimensions corresponding to \textit{sets} of entities,
so we have $2^N$~dimensions for $N$~entities.

\citet{copestake2012ideal} propose that dimensions could correspond
to logical expressions being true of an entity in a situation.
However, this requires generalising from an \textit{actual} distribution
(based on observed utterances)
to an \textit{ideal} distribution
(based on truth of logical expressions).
They do not propose a concrete algorithm,
but they discuss several challenges,
and suggest that grounded data might be necessary.
In this vein, \citet{kuzmenko2019real}
use the Visual Genome dataset \citep{krishna2017vg}
to learn vector representations with logically interpretable dimensions,
although these vectors are not as expressive as
\citeauthor{copestake2012ideal}'s ideal distributions.

Finally, a third option is to learn logical representations instead of vectors.
For example, in my own work I have represented words
as truth-conditional functions that are compatible with first-order logic
\citep{emerson2017b,emerson2020quant}.
Since referents are not observed in distributional semantics,
this introduces latent variables that make the model computationally expensive,
although there are ways to mitigate this \citep{emerson2020pixie}.


Despite the computational challenges,
I believe the right approach is to learn a logically interpretable model,
either by defining a vector space with logical structure,
or by directly using logical representations.
However, an important question is what kind of logic to use.
I argued in~\cref{sec:lex:vague}
that probabilities of truth and fuzzy truth values can capture vagueness,
and there are corresponding logics.

In probability logic, propositions have probabilities of being true or false,
with a joint distribution for the truth values of all propositions
\citex{for an introduction, see}{adams1998prob,demey2013prob}.
In fuzzy logic, propositions have fuzzy truth values,
and classical logical operators (such as:~$\wedge$,~$\vee$,~$\neg$)
are replaced with fuzzy versions
\citex{for an introduction, see}{hajek1998fuzzy,cintula2017fuzzy}.
Fuzzy operators act directly on truth values
-- for example, given the fuzzy truth values of $p$~and~$q$,
we can calculate the fuzzy truth value of~${p\vee q}$.
In contrast, in probability logic,
given probabilities of truth for $p$~and~$q$,
we cannot calculate the probability of truth for ${p\vee q}$,
unless we know the joint distribution.

A problem with fuzzy logic, observed by \citet{fine1975vague},
comes with propositions like~${p\vee \neg p}$.
For example, suppose we have a reddish orange object,
so the truth of \textit{red} and \textit{orange} are both below 1.
Intuitively, both \textit{red or not red} and \textit{red or orange}
should definitely be true.
However, in fuzzy logic, they could have truth below~1.
This makes probability logic more appealing than fuzzy logic.\footnote{%
	\citet{hajek1995logic} prove that fuzzy logic
	can be used to provide upper and lower bounds on probabilities in a probability logic,
	giving it a different motivation.
}

Furthermore, there are well-developed frameworks for probabilistic logical semantics
\citeg{goodman2015prob,cooper2015prob},
which a probabilistic distributional semantics could connect to,
or draw inspiration from.

%


\subsection{Context Dependence}
\label{sec:sentence:context}

The flipside of compositionality is \textit{context dependence}:
the meaning of an expression often depends on the context it occurs in.
For example, a \textit{small elephant} is not a \textit{small animal},
but a \textit{large mouse} is --
the meanings of \textit{small} and \textit{large}
depend on the nouns they modify.
One goal for a semantic model is to capture
how meaning depends on context.\footnote{%
	Ultimately, this must include dependence on real-world context.
	Even the intuitive conclusion that a large mouse is a small animal
	depends on the implicit assumption that you and I are both humans,
	or at least, human-sized.
	From the perspective of an ant,
	a mouse is large animal.
}

Following \citet{recanati2012composition}, we can distinguish
\textit{standing meaning}, the context-independent meaning of an expression,
and \textit{occasion meaning}, the context-dependent meaning
of an expression in a particular occasion of use.\footnote{%
	This terminology adapts \citet{quine1960meaning}.
}
However, every usage occurs in \textit{some} context,
so a standing meaning must be seen as an abstraction across usages,
rather than a usage in a ``null'' context
\citex{for discussion, see}{searle1980meaning,elman2009meaning}.
One approach is to treat a distributional vector as a standing meaning,
and modify it to produce occasion meanings.
For example, vectors could be modified according to syntactic or semantic dependencies
\citeg{erk2008context,thater2011context,dinu2012compare},
or even chains of dependencies \citeg{weir2016tree}.

This mapping from standing vectors to occasion vectors
can also be trained \citeg{czarnowska2019,popa2019context}.
Large language models such as ELMo \citep{peters2018elmo}
and BERT \citep{devlin2019bert}
can also be interpreted like this
-- these models map a sequence of input embeddings
to a sequence of contextualised embeddings,
which can be seen as standing meanings and occasion meanings, respectively.

Alternatively, standing meanings and occasion meanings
can be represented by different kinds of object.
\citet{erk2010exemplar} represent a standing meaning
as a set of vectors (each derived from a single sentence of the training corpus),
and an occasion meaning is a weighted sum of these vectors.

For a probabilistic model,
calculating an occasion meaning can be cast as Bayesian inference,
conditioning on the context.
This gives us a well-understood theoretical framework,
making it easier to generalise a model to other kinds of context.

\citet{dinu2010context} interpret a vector as a distribution over latent senses,
where each component is the probability of a sense.
Given probabilities of generating context words from latent senses,
we can then condition the standing distribution on the context.
However this model relies on a finite sense inventory,
which I argued against in~\cref{sec:lex:poly}.

\citet{lui2012usage} and \citet{lau2012sense,lau2014sense} use LDA \citep{blei2003lda},
where an occasion meaning is a distribution over context words
(varying continuously as topic mixtures),
and a standing meaning is a prior over such distributions.\footnote{%
	There are two distinct uses of a distribution here:
	to represent uncertainty, and to represent meaning.
	A sense is a topic mixture, parametrising a distribution over words;
	uncertainty is a Dirichlet distribution over topic mixtures.
}
A separate model is trained for each target word.
\citet{chang2014sense} add a generative layer,
allowing them to train a single model for all target words.
However, a single sense is chosen in each context,
giving a finite sense inventory. 

Skip-gram can be interpreted as
generating context words from a target word.
While we can see an embedding as a standing meaning,
nothing can be seen as an occasion meaning.
\citet{brazinskas2018bayes} add a generative layer,
generating a latent vector from the target word,
then generating context words from this vector.
We can see a latent vector as an occasion meaning,
and a word's distribution over latent vectors as a standing meaning.

Finally, in my own work,
I have also calculated occasion meanings by conditioning on the context
\citep{emerson2017b},
but in contrast to the above approaches,
standing meanings are truth-conditional functions (binary classifiers),
which I have argued for elsewhere in this paper.

\section{Conclusion}
\label{sec:conc}

A common thread among all of the above sections
is that reaching our semantic goals
requires structure beyond representing meaning as a point in space.
In particular, it seems desirable
to represent the meaning of a word as a region of space or as a classifier,
and to work with probability logic.

However, there is a trade-off between expressiveness and learnability:
the more structure we add,
the more difficult it can be to work with our representations.
To this end, there are promising neural architectures for working with structured data,
such dependency graphs
\citeg{marcheggiani2017graph}
or logical propositions
\citeg{rocktaeschel2017proving,minervini2018proving}.
To mitigate computationally expensive calculations in probabilistic models,
there are promising new techniques
such as amortised variational inference,
used in the Variational Autoencoder
\citep{kingma2013vae,rezende2014vae,titsias2014vae}.

My own recent work in this direction has been to develop
the Pixie Autoencoder \citep{emerson2020pixie},
and I look forward to seeing alternative approaches from other authors,
as the field of distributional semantics continues to grow.
I hope that this survey paper will help other researchers to develop the field
in a way that keeps long-term goals in mind.

\section*{Acknowledgements}

This paper is based on chapter 2 of my PhD thesis \citep{emerson2018}.
For invaluable advice on the structure and framing of that chapter,
and therefore also of this paper,
I want to thank my PhD supervisor Ann Copestake.
I would also like to thank my PhD examiners,
Katrin Erk and Paula Buttery,
for feedback on that chapter,
as well as Emily M.\ Bender, Guy Aglionby, Andrew Caines,
and the NLIP reading group in Cambridge,
for feedback on earlier drafts of this paper.
I would like to thank ACL reviewers 1 \& 3
for pointing out areas that were unclear,
and reviewer 2 for their kind praise.

I am supported by a Research Fellowship at Gonville \& Caius College, Cambridge.

\bibliography{thesis,pixie}
\bibliographystyle{acl_natbib}


\end{document}